\documentclass[sigconf,nonacm]{acmart}

\AtBeginDocument{%
}

\renewcommand\footnotetextcopyrightpermission[1]{}

\usepackage{algorithm}
\usepackage{algorithmic}
\usepackage{multirow}
\usepackage{enumitem}
\usepackage{seqsplit}

\begin{document}


\title[Gene Expression--Informed Generative Modeling]%
{Gene Expression--Informed Jointly Controlled Generative Modeling
for Precision Molecular Design}


\author{Hang Yuan}
\orcid{0009-0001-3135-582X}
\affiliation{%
  \department{School of Artificial Intelligence}
  \institution{South China Normal University}
  \city{Foshan}
  \state{Guangdong}
  \country{China}
}
\affiliation{%
  \department{D3 Center}
  \institution{The University of Osaka}
  \city{Suita}
  \state{Osaka}
  \country{Japan}
}

\author{Chen Li}
\orcid{0000-0002-8784-8148}
\authornote{Chen Li and Yuncheng Jiang are the corresponding authors.}
\affiliation{%
  \department{D3 Center}
  \institution{The University of Osaka}
  \city{Suita}
  \state{Osaka}
  \country{Japan}
}

\author{Wenjun Ma}
\orcid{0000-0003-4600-398X}
\affiliation{%
  \department{School of Artificial Intelligence}
  \institution{South China Normal University}
  \city{Foshan}
  \state{Guangdong}
  \country{China}
}
\affiliation{%
  \department{School of Computer Science}
  \institution{South China Normal University}
  \city{Guangzhou}
  \state{Guangdong}
  \country{China}
}

\author{Tadahiko Murata}
\orcid{0000-0002-1654-3945}
\affiliation{%
  \department{D3 Center}
  \institution{The University of Osaka}
  \city{Suita}
  \state{Osaka}
  \country{Japan}
}

\author{Yuncheng Jiang}
\orcid{0000-0002-0402-5382}
\authornotemark[1]
\affiliation{%
  \department{School of Artificial Intelligence}
  \institution{South China Normal University}
  \city{Foshan}
  \state{Guangdong}
  \country{China}
}
\affiliation{%
  \department{School of Computer Science}
  \institution{South China Normal University}
  \city{Guangzhou}
  \state{Guangdong}
  \country{China}
}

\renewcommand{\shortauthors}{Yuan et al.}


% ...


\begin{abstract}
Precision molecular design aims to discover personalized drug candidates through joint control of multiple conditions, such as biological relevance and molecular design strategies. Biological relevance reflects cellular functional states under disease or perturbation conditions, while molecular design strategies provide complementary guidance in terms of structural intentions and property optimization. In this study, we propose \textbf{JoPMol}, a \textbf{\underline{jo}}intly controlled \textbf{\underline{p}}recision \textbf{\underline{mol}}ecular generative model that integrates biological states encoded by gene expression profiles with molecular structure information expressed in text, and chemical properties quantified by numerical values within a unified modeling framework. This formulation enables coordinated generation and optimization of candidate molecules under joint condition control. Experimental results show that JoPMol outperforms state-of-the-art methods across multiple evaluation metrics. Moreover, JoPMol demonstrates strong generalization ability in both transfer tasks and biologically grounded simulation scenarios, validating its effectiveness for precision molecular design. The source code is publicly available at \url{https://github.com/hala-yh/JoPMol}.
\end{abstract}






\maketitle

\section{Introduction}

Precision molecular design has attracted increasing attention in personalized drug discovery, aiming to generate candidate molecules that are both biologically relevant and chemically feasible under multiple design constraints \cite{zeng2022deep}. In practice, the effectiveness of a drug candidate depends on several complementary factors, including its compatibility with biological systems \cite{li2025ai,kaitoh2021triomphe}, its structural information \cite{edwards-etal-2022-translation,gong2024text}, and its chemical properties \cite{pathak2025ai,inukai2025leveraging}. Capturing these diverse aspects typically requires information from multiple heterogeneous sources that describe different facets of the drug discovery process \cite{zhou2025multimodal}. Effectively integrating such heterogeneous information remains a challenge for precision molecular design.

From a biological perspective, gene expression profiles provide valuable information about disease states and cellular responses to drug perturbations, offering biologically grounded guidance for precision molecular design \cite{matsukiyo2024transcriptionally,loeffler2024reinvent}. These genetic signatures capture complex cellular phenotypes and have increasingly been used to inform drug discovery and molecular design \cite{kaitoh2021triomphe}. However, gene expression profiles inherently exhibit high dimensionality, substantial noise, and complex inter-gene dependencies without explicit structural correspondence to molecules \cite{li2025ai}. Moreover, expression patterns vary significantly across disease types and individual patients, further increasing variability in the underlying biological signals \cite{kang2025deep}. These characteristics make it challenging to extract reliable and structurally relevant features for guiding molecular design when relying solely on gene expression profiles.

Beyond biological context, precision molecular design also needs to consider structural design intentions \cite{wang2019smiles} and chemical property optimization \cite{liu2018constrained}. One line of research approaches this problem through text-driven molecular design, where natural language descriptions are used to express structural requirements such as functional groups or molecular structures \cite{luo2024text,jin2018junction}. These descriptions provide an interpretable interface that allows researchers to communicate design intentions in a flexible and human-readable manner \cite{weng2025text}. However, structural design intentions are typically defined at the molecular level and lack explicit alignment with disease-specific biological contexts, limiting their ability to provide precise guidance for therapeutic objectives \cite{edwards-etal-2022-translation}. Consequently, text-driven approaches tend to focus on satisfying structural constraints, while overlooking the optimization of biologically relevant and pharmacologically important properties. Quantitative properties such as solubility, synthesizability, and drug-likeness provide measurable optimization objectives that encourage generated molecules to satisfy practical design criteria \cite{khater2025generative}. Property-guided approaches therefore enable models to explore chemically meaningful regions of the molecular design space \cite{kong2024molecule,igashov2024equivariant}. Nevertheless, optimizing numerical properties alone often focuses on theoretical objectives defined in chemical space, without incorporating biological information from real cellular systems. As a result, designed molecules may satisfy predefined property constraints but remain disconnected from the biological contexts in which drugs ultimately function. 

These signals indicate the importance of integrating multiple sources of information for precision molecular design. However, most existing molecular design frameworks rely on a single source of information or only partially incorporate multiple design conditions \cite{moret2020generative}. Biological guidance, structural design intentions, and chemical properties originate from heterogeneous modalities with distinct semantic roles and representation spaces, and are therefore often treated in isolation. Such practices fail to capture their intrinsic dependencies and complementary effects, leading to suboptimal control over molecular design \cite{wang2026multimodal}. In particular, to the best of our knowledge, there is currently no molecular design framework that jointly integrates gene expression profiles, structural design intentions, and chemical properties within a unified generative model. Two key challenges contribute to this limitation. First, these modalities are typically collected from independent data sources, lacking unified and well-aligned datasets that jointly capture biological signals, structural intentions, and chemical properties. Second, they differ fundamentally in representation spaces and semantic granularities, which makes their coordinated modeling within a unified generative framework inherently challenging.

To address these limitations, we propose \textbf{JoPMol}, a \textbf{\underline{jo}}intly controlled \textbf{\underline{p}}recision \textbf{\underline{mol}}ecular generative model for precision molecular design. JoPMol jointly conditions molecular design on three complementary sources of information: biological states encoded by gene expression profiles, structural design intentions expressed in text, and chemical optimization objectives specified by numerical property values. To effectively integrate these heterogeneous signals, JoPMol employs source-specific feature extractors to learn modality-aware representations and introduces a bidirectional cross-attention mechanism to enable cross-modal interaction and information fusion. The resulting unified representation is used to guide a diffusion-based molecular generator, enabling end-to-end molecular design and optimization under multiple design constraints. The main contributions are summarized as follows:

\begin{itemize}[leftmargin=1em]
\item \textbf{Jointly controlled precision molecular design}:  
We introduce a jointly controlled generation approach under a multi-condition setting, enabling precision molecular design that simultaneously accounts for biological relevance, structural design intentions, and property optimization targets.

\item \textbf{Integration of design and optimization objectives}: 
We propose JoPMol, which learns latent representations of heterogeneous control signals and incorporates chemical properties as optimization objectives, enabling jointly controlled molecular design and optimization in a unified framework.

\item \textbf{Superior performance for precision molecular design}:  
JoPMol consistently outperforms state-of-the-art (SOTA) baselines across multiple evaluation metrics and diverse experimental settings, demonstrating its effectiveness and robustness for precision molecular design.
\end{itemize}

\section{Related Works}
\subsection{Gene-Guided Molecular Design}
Gene expression profiles characterize cellular responses to molecular perturbations, providing biologically relevant signals for molecular design \cite{matsukiyo2024transcriptionally}. 
These profiles provide a biologically grounded representation of disease states, making them a promising yet challenging signal for guiding molecular design in practice.
TRIOMPHE \cite{kaitoh2021triomphe} retrieves proxy molecules associated with gene expression profiles that are most similar to a target profile and uses them as inputs to a variational autoencoder (VAE) \cite{pinheiro2021variational} to generate molecules. Such profiles do not directly participate in the generation process, and similarity-based retrieval provides only a coarse approximation of complex biological conditions. In contrast, Gx2Mol \cite{li2024gx2mol} establishes an end-to-end mapping between genes and molecular structures, while GxVAEs \cite{li2025ai} further extend this line of work by jointly modeling gene expression profiles and inducing molecules using dual VAEs. Although these methods demonstrate the feasibility of leveraging biological signals for molecular design, they still treat gene expression profiles as the sole conditioning input and lack explicit mechanisms to jointly integrate structural design intentions and chemical property optimization within a unified generative framework \cite{guan2024drug}. As a result, their ability to support coordinated optimization under joint condition control remains limited in precision molecular design. 

\begin{figure*}[t]
\centering
\includegraphics[width=1.0\textwidth]{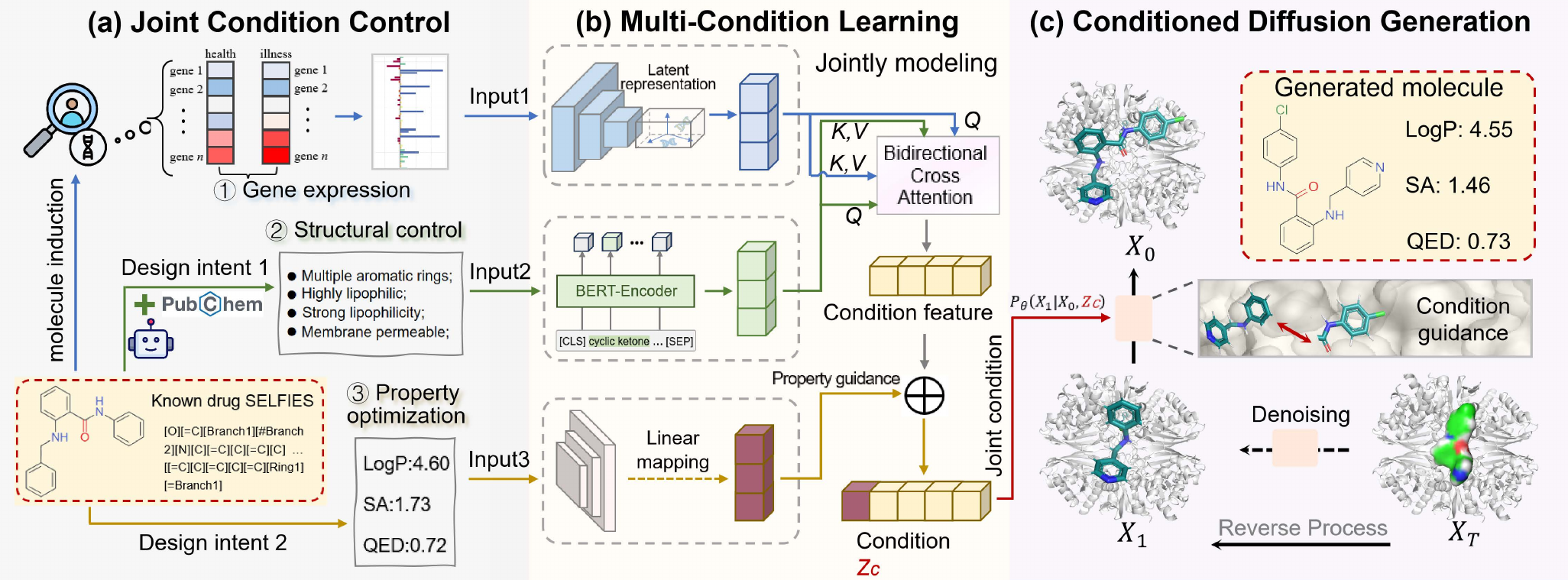}

\caption{Overview of the jointly controlled molecular generation framework for precision molecular design.
(a) \textbf{Joint condition control:} Three jointly controlled conditions are extracted, including gene expression profiles, textual structural intentions, and numerical property values. These complementary signals characterize biological relevance and molecular design strategies.
(b) \textbf{Multi-condition learning:} Biological and structural representations are used, and then the resulting features are further integrated with chemical property embeddings to form a joint conditional representation.
(c) \textbf{Conditioned diffusion generation.} The joint control embedding $\mathbf{z}_c$ guides the reverse diffusion process to progressively refine molecular representations, enabling precise molecular design and property optimization.}

\vspace*{-1.0\baselineskip} 
\label{fig1}

\end{figure*}

\subsection{Text-Based Molecular Design}
Text-based structural intentions provide an interpretable interface for specifying structural patterns of molecules, enabling controllable molecular generation beyond purely data-driven representations in a flexible and human-interpretable manner~\cite{deng2025chemical}. MolT5~\cite{edwards-etal-2022-translation} adopts a Transformer-based architecture to map natural language descriptions into molecular sequences. Notably, its autoregressive decoding paradigm often suffers from error sequence sorting and limited controllability over fine-grained structural modeling. Text+ChemT5~\cite{christofidellis2023unifying} enhances modeling capability through a multitask framework that jointly integrates chemical and natural language representations, but it remains constrained by autoregressive generation, making it difficult to balance semantic fidelity and structural consistency. TGM-DLM~\cite{gong2024text} adopts a diffusion-based framework to avoid the limitations of autoregressive generation. It realizes text-based molecular design through iterative denoising and conditional guidance, and further applies a nearest-neighbor projection to map continuous latent representations onto discrete molecular tokens, enabling the recovery of molecular sequences. Such methods show the feasibility of using natural language descriptions to express structural design intentions for precision molecular design. However, most existing text-guided models do not explicitly coordinate structural intentions with physicochemical property optimization or biological context, which limits their ability to support coherent and stable control under multiple design objectives.

\subsection{Property-Oriented Molecular Design}
Explicit numerical property values provide direct and quantitative optimization targets for molecular design, enabling fine-grained control over properties in a controllable and interpretable manner~\cite{hou2025novo}. MolSearch~\cite{sun2022molsearch} formulates the generation as a neural search process, iteratively producing and scoring candidate molecules to refine property optimization trajectories, but relies on heuristic scoring and does not preserve molecular semantic structures during optimization. 
DyMol~\cite{shin2024dynamic} adopts a reinforcement learning algorithm that updates policies based on oracle feedback to optimize properties under conditions, while the learning process remains sensitive to reward design and exhibits limited structural controllability. 
FRATTVAE~\cite{inukai2025leveraging} incorporates structure-aware attention into VAEs to improve property-controllable design. Nevertheless, property constraints are not aligned with structural semantics, which limits fine-grained structural control~\cite{angelo2023multi}. Moreover, such methods focus on single properties or limited property combinations and struggle to effectively coordinate with complex biological signals or design strategies, limiting their application in multi-condition molecular design. In this study, we establish jointly controlled conditions by associating gene expression profiles with structural design intentions, and incorporate numerical property values as explicit optimization objectives into the precision molecular design framework.

\section{Methodology}

\subsection{Problem Definition}

Precision molecular design aims to generate candidate molecules that satisfy multiple objectives under heterogeneous conditions, including biological relevance, structural intentions, and chemical property requirements.
We formalize these heterogeneous design conditions as three types of inputs. Specifically, biological relevance is represented by a gene expression profile
$\mathbf{G} = [g_1, g_2, \ldots, g_K] \in \mathbb{R}^K$, 
where $K$ denotes the number of genes and $g_k$ represents the expression value of the $k$-th gene. Structural intentions are represented as a textual sequence
$\mathbf{S} = [s_1, s_2, \ldots, s_N]$, 
where $N$ denotes the sequence length and each $s_n$ corresponds to a token describing molecular design intent. Chemical property requirements are represented as a property vector
$\mathbf{P} = [p_1, p_2, \ldots, p_I] \in \mathbb{R}^I$, 
where $I$ denotes the number of selected properties and $p_i$ represents a numerical chemical property.
The target is a molecular sequence $\mathbf{X} = [x_1, x_2, \ldots, x_M]$, represented in SELFIES, where $M$ denotes the maximum sequence length and $x_m$ is a discrete token.
The objective is to learn a conditional generative model $p_{\theta}(\mathbf{X} \mid \mathbf{G}, \mathbf{S}, \mathbf{P})$, which enables controllable molecular generation under jointly conditioned heterogeneous modalities, including gene expression profiles, structural design intentions, and chemical property constraints.

\textbf{Theory Motivation}:
Recent studies in complex systems and representation learning suggest that effective modeling of complex phenomena requires decomposing the system into modules with distinct functional roles \cite{battaglia2018relational,goyal2022inductive}. Instead of forcing heterogeneous signals into a unified representation, a modular and compositional perspective advocates modeling different factors separately according to their roles in the generation process.
We adopt this perspective for precision molecular design. Gene expression profiles $\mathbf{G}$ and structural design intent $\mathbf{S}$ provide rich semantic information that constrains the space of feasible molecular structures by specifying biological context and structural requirements. In contrast, chemical properties $\mathbf{P}$ exhibit a many-to-one relationship with molecular structures, where diverse molecules can share similar property values. As a result, property signals alone are insufficient to determine structure, and instead act as constraints that bias the generation toward desired objectives.
This functional distinction leads to a structured formulation:
\begin{equation}
p(\mathbf{X} \mid \mathbf{G}, \mathbf{S}, \mathbf{P}) = p(\mathbf{X} \mid \mathbf{Z}_{\text{sem}}, \mathbf{P}), \quad \mathbf{Z}_{\text{sem}} = f(\mathbf{G}, \mathbf{S}),
\end{equation}
where $f(\cdot)$ denotes a learnable mapping that integrates gene expression profiles and structural intent into a unified semantic representation, and $\mathbf{Z}_{\text{sem}}$ captures the resulting structural semantics under biological context. The property vector $\mathbf{P}$ modulates the generation process as a separate conditioning signal without directly determining the structure.

Following this observation, JoPMol adopts a hierarchical multimodal fusion strategy, where the aligned features of gene expression profiles and structural design intentions $\mathbf{Z}_{\text{sem}}$ are first integrated to construct semantic information, and property signals are introduced as optimization objectives to guide the conditional diffusion process, enabling jointly controlled molecular design under heterogeneous conditions. Figure~\ref{fig1} illustrates the JoPMol framework designed based on this principle, consisting of three key components. The joint condition control module parses and organizes jointly controlled conditions. The multi-condition learning module performs representation learning over multiple control signals and constructs unified conditional representations. The conditioned diffusion generation module enables precise molecular design and property optimization under joint control.

\subsection{Joint Condition Modeling}

We formulate molecular generation under joint control by constructing a unified condition tuple $\mathcal{C} = (\mathbf{G}, \mathbf{S}, \mathbf{P}, \mathbf{X})$, which integrates biological response signals, structural design intent, and property optimization targets. Given an induced molecule, its perturbation elicits transcriptional responses in cells, resulting in a differential gene expression profile $\mathbf{G}$ that reflects the underlying biological state. The molecule itself is represented as a SELFIES sequence $\mathbf{X}$ \cite{krenn2020self}. To associate biological signals with explicit design strategies, we derive structural intentions and optimization objectives from the molecule. A structural design intent $\mathbf{S}$ is generated using BioT5 \cite{pei-etal-2023-biot5}, adapted using the PubChem database \cite{kim2016pubchem}, to express structural patterns and drug-like characteristics. All generated structural descriptions are manually reviewed, with detailed annotation procedures provided in Appendix \ref{appendix_B}. In addition, a set of quantitative chemical properties $\mathbf{P}$ is computed to provide interpretable optimization targets for molecular design.
This joint condition formulation establishes a coherent connection between biological response, structural semantics, and property constraints, enabling controlled molecular design under heterogeneous conditions.
\subsection{Multi-Condition Learning}
We realize the $\mathbf{Z}_{\text{sem}} = f(\mathbf{G}, \mathbf{S})$ by learning a unified representation that integrates biological relevant and structural intent, while treating chemical properties $\mathbf{P}$ as separate conditioning signals.

\textbf{Biological Representation Learning.}
Gene expression profiles exhibit high dimensionality and substantial noise, making it challenging to directly employ them as conditioning signals for generative modeling \cite{li2025ai}. We adopt a VAE to learn a compact and smooth biological latent representation. The VAE encodes $\mathbf{G}$ into a latent variable $\mathbf{z}_g$, which captures essential transcriptomic patterns while filtering out spurious variations. By enforcing a continuous and approximately Gaussian latent manifold, the VAE provides well-behaved conditioning signals that are compatible with diffusion-based generative modeling and improve sampling stability under heterogeneous biological conditions \cite{venkatramanoutsourced}.
The VAE encoder is trained by minimizing the evidence lower bound:
{\small
\begin{equation}
\mathcal{L}_{\text{G}} =
\mathbb{E}_{q_{\phi}(\mathbf{z}_g \mid \mathbf{G})}
\left[
\left\| \hat{\mathbf{G}} - \mathbf{G} \right\|_2^2
\right]
+ \beta \, \mathrm{KL}
\left(
q_{\phi}(\mathbf{z}_g \mid \mathbf{G}) \,\|\, p(\mathbf{z}_g)
\right),
\end{equation}}
where $q_{\phi}(\cdot)$ denotes the encoder posterior distribution, $\hat{\mathbf{G}}$ is the reconstructed gene expression, $p(\cdot)$ is the prior distribution, and $\beta$ controls the strength of latent regularization. 
The learned latent variable $\mathbf{z}_g$ provides a continuous and biological embedding space that facilitates interpolation and generalization across heterogeneous perturbation conditions. For compatibility with token-based attention mechanisms, $\mathbf{z}_g$ is further reshaped and projected into a token-level biological embedding 
$\mathbf{Z}_g \in \mathbb{R}^{T_g \times D}$, 
where $T_g$ represents the number of tokens, and $D$ denotes the embedding dimension.

\textbf{Structural Design Intention Representation Learning.}
Structural design intentions are derived from the induced molecule and expressed as semantic descriptions that encode molecular structures. 
We employ BioLink-BERT \cite{yasunaga2022linkbert} to transform the structural semantic representation $\mathbf{S}$ into a contextualized embedding 
$\mathbf{Z}_s = f_{S}(\mathbf{S}) \in \mathbb{R}^{T_s \times D}$, 
where $f_{S}(\cdot)$ denotes the BioLink-BERT encoder, $T_s$ represents the number of structural tokens. 
The resulting embedding $\mathbf{Z}_s$ provides interpretable structural guidance that complements the latent biological representation.

\textbf{Biological–Structural Semantic Alignment.}
To construct the unified semantic representation $\mathbf{Z}_{\text{sem}} = f(\mathbf{G}, \mathbf{S})$, we employ a bi-directional cross-attention mechanism that enables reciprocal information exchange between $\mathbf{Z}_g$ and $\mathbf{Z}_s$. Specifically, biological features attend to structural semantics as $\mathbf{Z}_g^{\prime} = \mathbf{Z}_g + \mathbf{CA}(\mathbf{Z}_g, \mathbf{Z}_s, \mathbf{Z}_s)$, allowing gene-level representations to absorb structurally relevant information, where $\mathbf{CA}(\cdot)$ denotes the multi-head cross-attention operator. Structural features attend to biological representations as $\mathbf{Z}_s^{\prime} = \mathbf{Z}_s + \mathbf{CA}(\mathbf{Z}_s, \mathbf{Z}_g, \mathbf{Z}_g)$, enabling structural tokens to become aware of underlying biological contexts. This reciprocal interaction promotes mutual calibration between conditions.

\textbf{Joint Conditional Representation Construction.}
Following the distinction between structural infomation and property constraints, we decouple semantic representation learning from property conditioning. After semantic alignment, biological and structural features are aggregated and normalized to form a unified semantic context. Rather than adopting an early-fusion strategy that entangles heterogeneous modalities, we preserve the semantic hierarchy by isolating structural representation from quantitative objective control. Chemical properties $\mathbf{P}$ are projected into the latent space via a linear mapping $f_{\text{proj}}(\cdot)$ to produce property control tokens, which are incorporated as prefix conditioning signals. This design enables explicit and stable regulation of molecular properties without disrupting the underlying semantic structure. The joint conditional representation is constructed as:
\begin{equation}
\mathbf{Z}_{\text{c}} =
\Big[
f_{\text{proj}}(\mathbf{P})
\;\oplus\;
\mathrm{LN}
\big(
\mathbf{Z}_g^{\prime} \oplus \mathbf{Z}_s^{\prime}
\big)
\Big],
\label{eq:joint_condition}
\end{equation}
where $\oplus$ denotes concatenation along the token dimension and $\mathrm{LN}(\cdot)$ denotes layer normalization. The resulting $\mathbf{Z}_{\text{c}}$ serves as the unified conditioning representation for the subsequent diffusion-based molecular generator, enabling integrated design and property-oriented optimization under jointly controlled conditions.

\subsection{Jointly Controlled Generative Modeling}
We embed the SELFIES sequence $\mathbf{X}$ of the molecule into a continuous latent space and perform diffusion on the resulting embeddings, enabling modeling of discrete molecular sequences.
Specifically, the tokens are mapped into embeddings
$\mathbf{E}_0 = \mathrm{Emb}(\mathbf{X}) \in \mathbb{R}^{L \times D}$, where $L$ denotes the sequence length. Following the standard forward diffusion process, Gaussian noise is integrated as
\begin{equation}
\mathbf{E}_t = \sqrt{\bar{\alpha}_t}\mathbf{E}_0 + \sqrt{1-\bar{\alpha}_t}\boldsymbol{\epsilon}, 
\quad \boldsymbol{\epsilon} \sim \mathcal{N}(\mathbf{0},\mathbf{I}),
\end{equation}
where $\bar{\alpha}_t$ denotes the cumulative noise schedule.

JoPMol is trained to learn a jointly conditioned denoising network $\epsilon_\theta(\mathbf{E}_t, t; \mathbf{Z}_{\text{c}})$ 
to predict the noise, conditioned on the joint control representation $\mathbf{Z}_{\text{c}}$ and the diffusion timestep $t$. Let $\hat{\mathbf{E}}_0$ be the recovered clean embedding. The overall jointly controlled training objective is formulated as
\begin{equation}
\mathcal{L}_{} =
\mathbb{E}_{t,\boldsymbol{\epsilon}}\Big[
\|\boldsymbol{\epsilon} - \epsilon_\theta(\mathbf{E}_t, t; \mathbf{Z}_{\text{c}})\|_2^2
+ \mathcal{L}_{\text{con}}(\hat{\mathbf{E}}_0, \mathbf{X})
+ \mathcal{L}_T
+ \lambda \mathcal{L}_{\text{cos}}
\Big],
\end{equation}
where the first term corresponds to the standard denoising objective in diffusion models. $\mathcal{L}_{\text{con}}$ enforces discrete token consistency via vocabulary logits, and $\mathcal{L}_T$ stabilizes the terminal diffusion We further introduce a cosine alignment loss $\mathcal{L}_{\text{cos}}$ to encourage semantic consistency between the biological-related and structural-related representations in the shared semantic space. The weighting coefficient $\lambda$ controls the contribution of the alignment objective, balancing biological and structural signals in the model.
At $t=0$, we match the predicted embedding with the original embedding to further stabilize discrete reconstruction in JoPMol.

\section{Experiments}
\subsection{Experimental Setup}

\textbf{Datasets.} Three cancer-related datasets were used. Each dataset consists of a 978-dimensional gene expression profile, the corresponding molecule represented in SELFIES format, structural design intentions, and numerical chemical property values. Table~\ref{tab:exp_data} reports dataset statistics, with more details in Appendix \ref{appendix_A}.

\begin{table}[h]
\caption{Statistics of the datasets. MLen: average SELFIES length; GDim: gene expression dimensionality; TLen: average structural text length; MW: average molecular weight.
}
\centering
\resizebox{0.46\textwidth}{!}{
\begin{tabular}{l|cccccccc}\toprule
Dataset & MLen & GDim & TLen & MW & LogP & SA & QED \\\midrule
TrainSet & 254 & 978 & 85 & 432 & 2.81 & 3.56 & 0.59 \\
TestSet & 211 & 978 & 85 & 355 & 3.32 & 2.80 & 0.62 \\
TransferSet & 249 & 978 & 78 & 397 & 3.70 & 2.69 & 0.56 \\
DiseaseSet & 272 & 884 & 54 & 463 & 0.22 & 4.30 & 0.40 \\\bottomrule
\end{tabular}
}
\vspace*{-1\baselineskip} 
\label{tab:exp_data}
\end{table}

\begin{table*}[t]
\caption{Comparison of JoPMol with baseline methods across structure consistency, sequence similarity, property optimization, and overall statistics. Best results are highlighted in bold, and second-best results are underlined.}
\centering
\setlength{\tabcolsep}{4pt}
\resizebox{1.0\textwidth}{!}{
\begin{tabular}{l|ccc|ccc|c@{\hspace{12pt}}cc|cccc}\toprule
\multirow{2}{*}{Method} & \multicolumn{3}{c|}{Structure Consistency} & \multicolumn{3}{c|}{Sequence Similarity} & \multicolumn{3}{c|}{Property Optimization} & \multicolumn{4}{c}{Overall Statistics}\\
 & ECFP$\uparrow$ & APFP$\uparrow$ & Dice$\uparrow$ & BLEU$\uparrow$ & Leven$\downarrow$ & LCS$\uparrow$ & LogP & SA & QED & Diversity$\uparrow$ & Total$\uparrow$ & FCD$\downarrow$ & Rank$\downarrow$ \\
\midrule
TRIOMPHE \cite{kaitoh2021triomphe}      & 0.17 & 0.32 & 0.27 & 0.24 & 33.53 & 0.51 & 3.55 & 3.40 & 0.54 & 80.77 & 0.35 & 27.49 & 7.17 \\
Gx2Mol \cite{li2024gx2mol}        & 0.19 & 0.36 & 0.28 & 0.30 & 31.32 & 0.56 & 2.96 & 3.51 & \underline{0.59} & 85.63 & 0.55 & 26.41 & 5.83 \\
GxVAEs \cite{li2025ai}       & \underline{0.22} & 0.36 & 0.26 & 0.43 & 30.53 & 0.58 & 2.95 & 3.63 & \underline{0.59} & 84.73 & 0.68 & 25.54 & 5.42 \\
Molsearch \cite{sun2022molsearch}     & 0.12 & 0.26 & 0.21 & 0.14 & 39.41 & 0.32 & 4.68 & 3.29 & 0.47 & 86.23 & \underline{0.82} & 10.92 & 7.33 \\
DyMol \cite{shin2024dynamic}     & 0.11 & 0.24 & 0.20 & 0.13 & 38.99 & 0.32 & 2.44 & \textbf{3.09} & 0.73 & 87.54 & \underline{0.82} & 13.05 & 7.17 \\
FRATTVAE \cite{inukai2025leveraging}     & 0.11 & 0.25 & 0.19 & 0.28 & 42.03 & 0.34 & \textbf{3.34} & 3.23 & 0.58 & 88.62 & \textbf{0.93} & 14.80 & 6.08 \\
MolT5 \cite{edwards-etal-2022-translation}       & 0.21 & 0.38 & 0.29 & 0.47 & 39.52 & 0.59 & 2.87 & 3.94 & \underline{0.59} & 88.91 & 0.71 & \underline{8.54} & 4.58 \\
Text+ChemT5 \cite{christofidellis2023unifying}  & 0.20 & 0.38 & \underline{0.38} & \underline{0.50} & 42.11 & 0.56 & 3.54 & 4.00 & 0.50 & 87.80 & 0.63 & 10.48 & 5.50 \\
TGM-DLM \cite{gong2024text}      & \underline{0.22} & \underline{0.40} & 0.32 & \underline{0.50} & \underline{26.07} & \underline{0.60} & 2.88 & 3.77 & 0.58 & \textbf{89.99} & 0.71 & 8.94 & \underline{3.50} \\
\midrule
JoPMol       & \textbf{0.33} & \textbf{0.51} & \textbf{0.46} & \textbf{0.61} 
            & \textbf{19.34} & \textbf{0.69} & \underline{3.27} 
            & \underline{3.12} & \textbf{0.62} & \underline{89.04} & \textbf{0.93} & \textbf{7.40} & \textbf{1.25} \\
\bottomrule
\end{tabular}
}

\label{tab:exp1}
\vspace*{-1.0\baselineskip} 
\end{table*}


\begin{itemize}[leftmargin=1em]
\item \textbf{TrainSet and TestSet} are collected from a human breast cancer cell line under diverse chemical perturbations, comprising 13,755 small molecules \cite{duan2014lincs}, randomly split at a ratio of 9:1 for training and evaluation of JoPMol.

\item \textbf{TransferSet} constructed using data from the same breast cancer cell line under genetic perturbations, including gene overexpression (AKT1, AKT2, AURKB, CTSK, EGFR, HDAC1, MTOR, and PIK3CA) and gene knockdown (SMAD3 and TP53), used to analyze the structural consistency between designed molecules and target-associated ligands in transfer tasks.

\item \textbf{DiseaseSet} derived from the CREEDS database \cite{wang2016extraction}, constructed by averaging gene expression profiles from multiple gastric cancer patients, used as a disease-level case study for evaluating model performance in biologically grounded and clinically relevant simulation scenarios.

\end{itemize}

\textbf{Evaluation Measures.} We evaluate the quality of the designed molecules from four complementary perspectives, including molecular structure, molecular sequence representation, chemical properties, and statistical metrics. 
\begin{itemize}[leftmargin=1em]
\item \textbf{Structure consistency}: \textit{ECFP} (extended-connectivity fingerprints) is used to quantify the consistency of local substructures between designed molecules and the targets. \textit{APFP} (atom pair fingerprints) encodes atom-type pairs together with their topological distances, effectively capturing global structural characteristics. \textit{Dice} is employed to compute the similarity between two APFP representations for structural comparison.

\item \textbf{Sequence similarity:} \textit{BLEU} measures the n-gram overlap between produced SELFIES sequences and the target sequences. \textit{Leven} (Levenshtein) distance quantifies the minimum number of edit operations required to transform a generated SELFIES into the corresponding target sequence, while \textit{LCS} (longest common subsequence) evaluates overall sequential consistency. 

\item \textbf{Property optimization:} \textit{LogP} characterizes molecular lipophilicity, \textit{SA} reflects synthetic accessibility, and \textit{QED} measures overall drug-likeness. 
These properties are selected as they capture complementary aspects of molecular quality, including pharmacokinetic behavior, synthetic feasibility, and overall drug-likeness, and are widely adopted in molecular generation studies.
Greater agreement between the average property values of designed molecules and the corresponding test-set statistics indicates improved distributional consistency and controllability.

\item \textbf{Overall statistics:} \textit{Diversity} measures structural diversity among candidate molecules. \textit{Total} score integrates validity, uniqueness, and novelty, providing an overall measure of molecular design performance. \textit{FCD} (Fréchet ChemNet distance) quantifies the distributional similarity between designed molecules and the targets in the learned chemical feature space. \textit{Rank} is obtained by averaging each model’s ranking positions across all metrics, where a lower rank indicates better overall performance. Additional details are provided in Appendix~\ref{appendix_D}. 
\end{itemize}

\textbf{Implementation Details.}
JoPMol is implemented in PyTorch. The gene encoder adopts multi-layer feedforward architectures with a final latent embedding dimension of 64. Structural intentions are encoded into fixed-length semantic representations, while SELFIES tokens are embedded into a 32-dimensional trainable embedding space with a maximum sequence length of 256. The joint conditional diffusion is trained for 2,000 timesteps using a cosine noise schedule and optimized with AdamW at a learning rate of $1\times10^{-4}$ and a dropout rate of 0.1 for regularization. A uniform skip-sampling strategy is adopted during reverse diffusion to improve sampling efficiency and generation stability. 
We analyze the sensitivity of the weighting coefficient $\lambda$ in Table~\ref{table:metrics_l}. We observe that when $\lambda$ is set to 0.3, the model achieves the best performance in cross-modal alignment.
More details are provided in Appendix~\ref{appendix_E}.

\subsection{Overall Performance under Joint Control}

To the best of our knowledge, there is no unified molecular generation framework that jointly integrates gene expression profiles, structural design intentions, and chemical properties. Therefore, we select baseline models based on their input modalities, including gene-based, text-based, and property-guided approaches. Since JoPMol operates under a multi-condition setting, the comparison aims to evaluate the effectiveness of jointly modeling heterogeneous conditions rather than enforcing strict equivalence across inputs.

Table~\ref{tab:exp1} presents a comprehensive comparison between JoPMol and multiple baseline models. The results indicate that JoPMol outperforms all baselines on the six metrics related to structural consistency and sequence similarity. Among the six metrics associated with chemical property optimization and overall statistics, JoPMol achieves the best performance on three metrics and the second-best performance on the remaining three, demonstrating a clear overall advantage. More specifically, compared with SOTA models such as TGM and GxVAEs, JoPMol improves the ECFP score by approximately 50\%, indicating its ability to reliably preserve local chemical structures and design candidate molecules capable of inducing the target gene expression profiles. For property optimization, even under multi-condition control, JoPMol produces molecules whose property distributions closely match those of the test set using only numerical property values as conditioning signals, demonstrating effective property controllability and optimization. Additionally, the overall statistics indicate that JoPMol achieves high validity and novelty and attains the best overall rank among all compared methods. These results show that JoPMol exhibits stable design performance under jointly controlled settings.

Furthermore, we validate the biological fidelity of the learned gene representations by reconstructing the latent features through the VAE decoder and comparing them with the original gene expression profiles. The reconstructed profiles exhibit high consistency with the original ones in both distributional patterns and key expression characteristics, supporting the reliability and biological plausibility of the learned representations. Detailed quantitative results and visualizations are reported in Appendix Figure \ref{fig:VAE}.

We further evaluate the distributional quality of the designed molecules using RDKit and MACCS similarity metrics, as summarized in Appendix Table \ref{tab:exp1_res}. The results demonstrate that JoPMol achieves the best overall performance, indicating close alignment with the empirical molecular distribution in terms of structural coverage and feature overlap. 

\subsection{Transfer Tasks under Genetic Perturbations}

\begin{figure}[t]
\centering
\includegraphics[width=1.0\hsize]{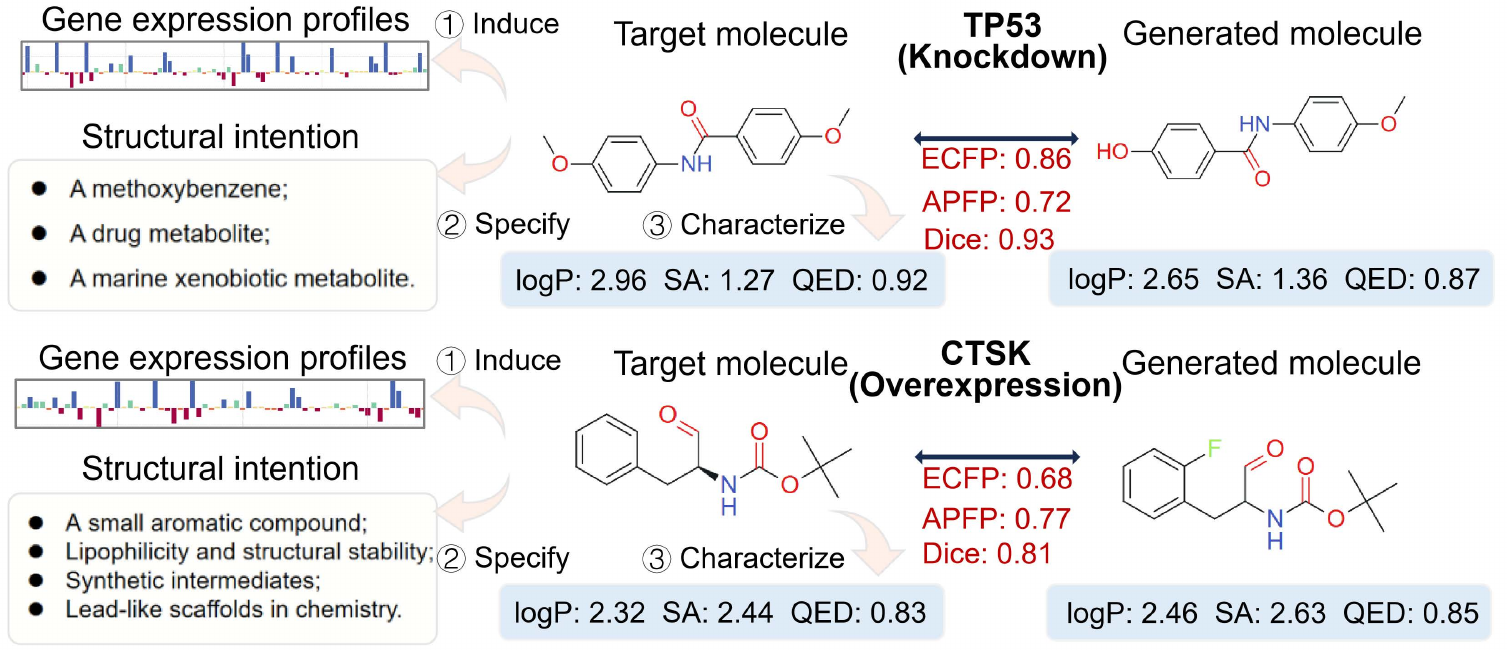}
\caption{Representative examples of multi-condition controlled molecular design under gene expression profiles, structural design intentions, and numerical property values for TP53 and CTSK.}
\label{fig3}
\vspace*{-1.0\baselineskip} 
\end{figure}

To further evaluate the generalization capability and transfer stability of JoPMol, we extend JoPMol to ten ligand conditions and assess its performance accordingly. Figure~\ref{fig3} presents two representative cases corresponding to gene knockdown (TP53) and gene overexpression (CTSK) scenarios. In each case, gene expression profiles induced by the target molecules, together with molecular design strategies, jointly guide the generation of candidate molecules, which are then compared against the target molecules.

In the gene knockdown scenario, the designed molecules preserve key functional-group patterns and achieve high structural similarity, with ECFP, APFP, and Dice scores of 0.86, 0.72, and 0.93, indicating that JoPMol effectively captures molecular structural characteristics under unseen genetic perturbations. Meanwhile, the three chemical properties of the candidate molecules remain highly consistent with those of the target molecules, demonstrating that JoPMol can effectively control chemical properties while preserving structural fidelity, enabling integrated molecular generation and optimization. Additionally, in the gene overexpression scenario, JoPMol continues to design molecules with reasonable consistency in structural patterns and local functional-group configurations, achieving ECFP, APFP, and Dice scores of 0.68, 0.77, and 0.81, while maintaining strong property controllability. Experimental results for the remaining eight transfer tasks are provided in Figure \ref{fig:additional_ligands}. 

Overall, JoPMol maintains both structural fidelity and property optimization capability under different types of genetic perturbations, demonstrating its effectiveness on transfer tasks and its potential for precision molecular design.

\subsection{Ablation Studies on Joint Controllability}
\begin{table}[t]
\caption{Ablation results at the overall statistics level under different fusion strategies and control settings. 
W/concat, W/cross, W/FiLM, and W/adapter denote variants that replace the multimodal fusion in JoPMol with concatenation, cross-attention, FiLM-based modulation, and Adapter-based fusion. 
W/gene, W/struct, and W/prop correspond to gene-controlled, structure-controlled, and property-controlled settings.}
\centering
\setlength{\tabcolsep}{2pt}
\resizebox{0.48\textwidth}{!}{
\begin{tabular}{l|cccccccc}\toprule
Setting & Total & Validity & Uniqueness & Novelty & Diversity & FCD & Rank \\
\midrule
W/concat    & 0.90 & 100.0 & 98.84 & 91.56 & 89.02 & 16.80 & 3.33 \\
W/cross     & 0.89 & 100.0 & 94.05 & 94.13 & 88.61 & 27.31 & 4.83 \\
W/FiLM     & 0.89 & 100.0 & 98.11 & 88.90 & 89.00 & 16.07 & 3.17 \\
W/adapter    & 0.84  & 100.0 & 90.07 & 94.24 & 85.05 & 23.91 & 6.83 \\
\midrule
W/gene    & 0.89 & 100.0 & 96.73 & 92.23 & 82.81 & 24.53 & 5.33 \\
W/struct       & 0.86 & 100.0 & 97.31 & 88.27 & \textbf{90.58} & 25.32 & 5.17 \\
W/prop       & 0.92 & 100.0 & \textbf{99.13} & 92.38 & 88.39 & 12.93 & \underline{2.67} \\
\midrule
JoPMol       & \textbf{0.93} & \textbf{100.0} & \underline{98.91} & \textbf{94.29} & \underline{89.04} & \textbf{7.40} & \textbf{1.33} \\
\bottomrule
\end{tabular}
}

\label{tab:exp2_res}
\vspace*{-1.0\baselineskip} 
\end{table}
To analyze the effectiveness of different fusion strategies and the necessity of joint controllability, we conduct an extended ablation study including both fusion variants (W/concat, W/cross, W/FiLM, and W/PropAdapter) and single-control settings (W/gene, W/struct, and W/prop). Table~\ref{tab:exp2_res} summarizes the results.

\textbf{Fusion Strategies}.
We examine different fusion strategies and analyze their impact on joint controllability. The concatenation strategy (W/concat) aggregates heterogeneous features without explicit interaction, achieving moderate performance (Total: 0.90, FCD: 16.80) but failing to capture cross-modal dependencies. Cross-attention-based fusion (W/cross) enables dynamic alignment between modalities and improves novelty to 94.13, but leads to a substantially higher FCD of 27.31, indicating degraded distributional consistency. 
Feature-wise linear modulation (FiLM)~\cite{perez2018film} (W/FiLM) incorporates properties through feature-wise transformations and achieves more balanced performance (FCD: 16.07), suggesting improved stability. W/adapter injects property conditions via a lightweight adapter module~\cite{houlsby2019parameter}, enhancing novelty to 94.24 but still resulting in a relatively high FCD of 23.91, as it mainly performs local modulation without global guidance. 
In contrast, JoPMol adopts a hierarchical fusion strategy by first jointly modeling biological and structural signals through bidirectional cross-modal interaction, and subsequently incorporating property conditions as global guidance. This enables stable integration of heterogeneous information and leads to improved performance in both structural consistency and distributional alignment.

\textbf{Control Settings}.
We then analyze single-control settings. The results show that JoPMol exhibits advantages under joint condition control and achieves improved overall performance. W/gene, which relies only on gene control, degrades noticeably in diversity and FCD. W/struct, which uses only textual structural intentions, achieves the highest diversity (90.58) but remains limited in Total and FCD. W/prop, which uses only numerical chemical property values, attains the best novelty and a relatively low FCD (12.93), but at the cost of reduced structural consistency and diversity. In contrast, JoPMol achieves the best performance on total, FCD, and rank, indicating a more stable balance between design quality and distributional consistency. Although its diversity is slightly lower than that of W/struct, this mainly results from the contraction of the feasible generation space under multiple control constraints.

\begin{figure}[t]
\centering
\includegraphics[width=0.5\textwidth]{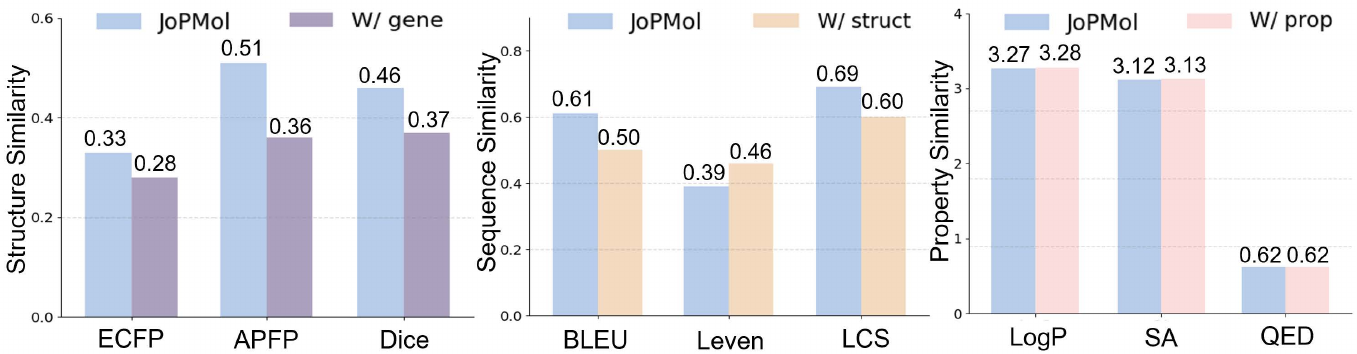}
\caption{Ablation results on joint controllability at different evaluation levels. Left: structure consistency; Middle: sequence similarity; Right: property optimization.}
\label{fig4}
\vspace*{-1\baselineskip} 
\end{figure}

Figure~\ref{fig4} further presents a comparison between joint condition control and single control in terms of structure, sequence, and property. For structural consistency, JoPMol achieves higher ECFP, APFP, and Dice scores than W/gene. For sequence similarity, JoPMol shows improvements in BLEU and LCS while substantially reducing the Leven distance. For property optimization, JoPMol performs comparably to W/prop on LogP, SA, and QED, with identical QED values, indicating strong alignment with the target distribution. Overall, joint condition control preserves property optimization without sacrificing structural consistency and sequential similarity, enabling precision molecular design.

Furthermore, we replace numerical chemical properties with textual property descriptions to investigate the impact of property representation on controllability. The results show that using textual descriptions leads to inferior overall performance compared with numerical values. Detailed results are provided in Table~\ref{table:F3}.

\subsection{Case Study I: Property Controllability}
\begin{figure*}[t]
\centering
\includegraphics[width=1.0\textwidth]{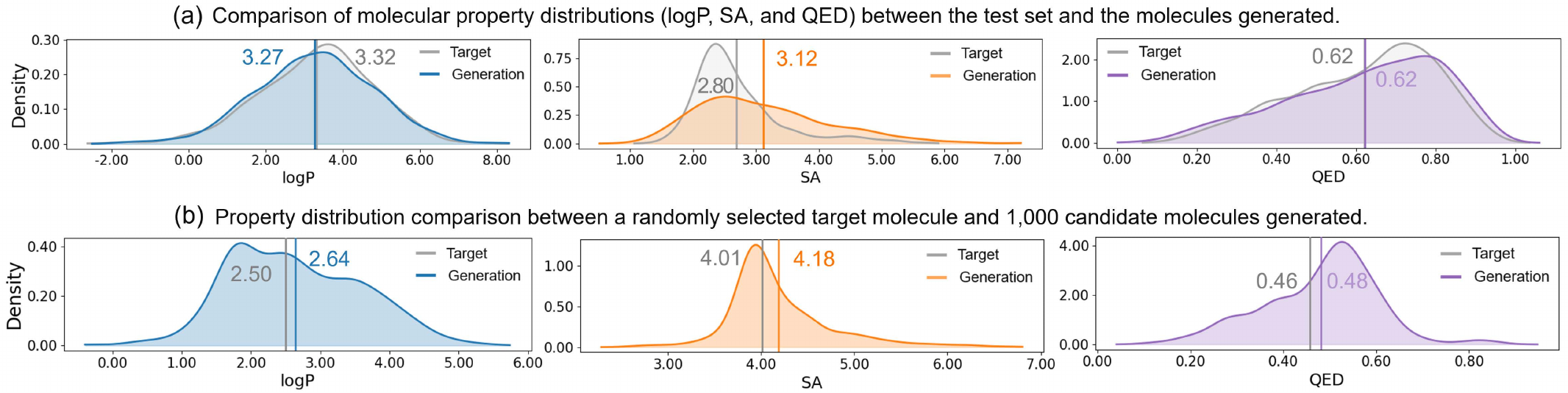}
\caption{Evaluation of molecular distribution alignment and property controllability achieved by JoPMol. (a) Distribution-level comparison of logP, SA, and QED between the test molecules and the candidate molecules. (b) Instance-level analysis comparing a randomly selected target molecule with 1,000 designed candidates.}
\vspace*{-1\baselineskip} 
\label{fig2}
\end{figure*}

To further evaluate the chemical property controllability of JoPMol, Figure~\ref{fig2} presents the distribution-level and instance-level results of the generated molecular properties. Figure~\ref{fig2}(a) compares the property distributions of the test-set molecules (gray curves) with those of the generated molecules (colored curves). For all three properties, the generated distributions closely match the target distributions in both overall shape and mean values, indicating that the model can stably capture and reproduce property targets at the population level without noticeable distribution shift or mode collapse. Additionally, the largest deviation is observed on SA, where the mean value shifts from 2.70 (target) to 3.12 (generated), corresponding to a relative deviation of approximately 15.6\%. This deviation is reasonable given the relatively wide value range of SA, while the deviations of logP and QED remain much smaller. These results indicate that JoPMol preserves the statistical structure of molecular property distributions while maintaining chemical plausibility.

Figure~\ref{fig2}(b) evaluates JoPMol at the instance level. Given a randomly selected target molecule, 1,000 candidate molecules are designed under its property constraints. The resulting property distributions are tightly concentrated around the target values, with mean values closely aligned with the specified conditions. The relative deviations between the designed means and the target values for logP, SA, and QED are all controlled within approximately 4-6\%, demonstrating accurate and stable conditional generation. 
Importantly, the narrow variance of the generated distributions suggests that JoPMol not only enforces property alignment but also reduces unnecessary structural dispersion. This behavior reflects the effectiveness of the proposed multimodal fusion and conditional diffusion mechanisms in guiding molecular design.

Overall, JoPMol achieves precise and reliable property controllability at both the distribution and instance levels, successfully balancing global distribution alignment with fine-grained, target-oriented optimization. This capability is critical for precision molecular design and provides a strong foundation for customizable molecular generation in downstream drug discovery applications.
\subsection{Case Study II: Biological Simulation}

\begin{figure}[t]
\centering
\includegraphics[width=0.48\textwidth]{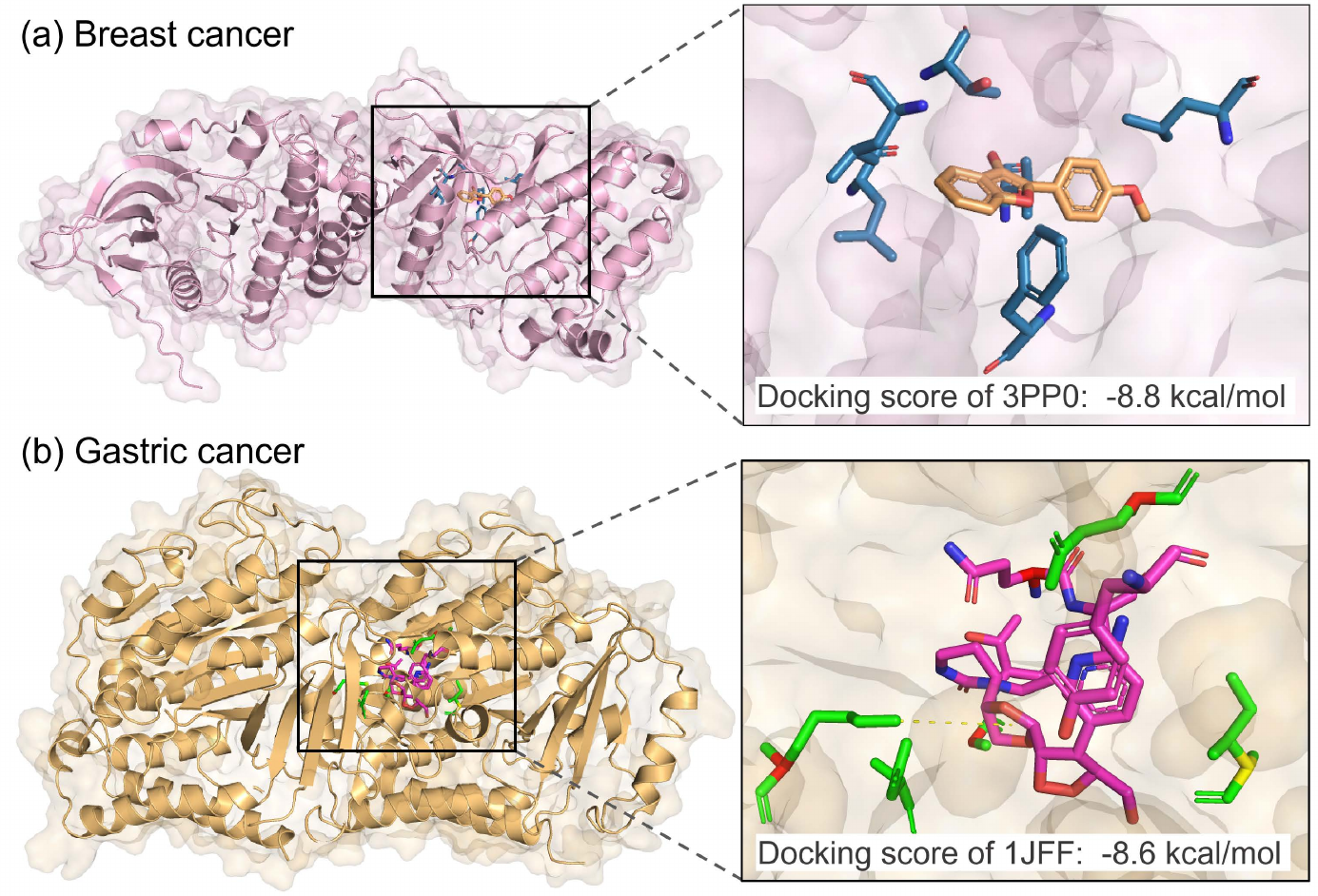}
\caption{Docking visualization of the designed molecules under (a) breast cancer and (b) gastric cancer settings. The predicted binding poses are shown within the binding pockets, with surrounding residues highlighted. The designed molecules achieve binding affinities of -8.8 $\mathrm{kcal/mol}$ on HER2 (PDB ID: 3PP0) and -8.6 $\mathrm{kcal/mol}$ on tubulin (PDB ID: 1JFF).}
\label{fig5}
\vspace*{-1.0\baselineskip} 
\end{figure}

To further evaluate the practical applicability and biological relevance of JoPMol, we conduct molecular docking studies under both breast cancer and gastric cancer disease settings.

\textbf{Breast cancer.} 
We perform molecular docking against a representative breast cancer target, HER2 (PDB ID: 3PP0), to evaluate the biological plausibility of the designed molecules within a disease-relevant context. As illustrated in Figure~\ref{fig5}(a), the designed molecule achieves a binding affinity of -8.8 $\mathrm{kcal/mol}$, outperforming a flavonoid-like reference compound \texttt{\seqsplit{COC1=CC=C(C=C1)C2=C(C(=O)C3=CC=CC=C3O2)OC}}, which obtains -8.4 $\mathrm{kcal/mol}$ under the same docking protocol. 
The enlarged view further shows that the designed molecule is well accommodated within the binding pocket and exhibits favorable spatial complementarity with surrounding residues. Specifically, the aromatic scaffold is positioned within the hydrophobic region of the pocket, enabling favorable van der Waals interactions, while the polar functional groups orient toward the pocket periphery, forming plausible hydrogen bonding interactions. Importantly, no obvious steric clashes or unrealistic conformations are observed, indicating a chemically plausible binding mode.

\textbf{Gastric cancer.} To further assess the generalization ability of JoPMol beyond the training distribution, we conduct a case study in a gastric cancer setting. Tubulin (PDB ID: 1JFF), a canonical target of taxanes widely used in gastric cancer therapy~\cite{ma2023efficacy}, is selected as the receptor. Molecular docking is performed to evaluate the binding behavior of the designed molecules in a cross-disease scenario.
As illustrated in Figure~\ref{fig5}(b), the designed molecule occupies the binding pocket of the target protein and exhibits reasonable spatial complementarity with surrounding amino acid residues. The docking result indicates a predicted binding energy of -8.6 $\mathrm{kcal/mol}$ on the 1JFF target, suggesting favorable binding potential. The enlarged view further visualizes the spatial interactions between the ligand and key residues, indicating a plausible binding conformation without apparent steric clashes or unstable orientations. 

These results collectively demonstrate that JoPMol not only produces biologically meaningful molecules consistent with the training domain, but also exhibits promising generalization capability to unseen disease contexts, highlighting its potential for cross-disease precision molecular design.

\section{Conclusion}
In this study, we propose JoPMol, a jointly controlled molecular generative model for precision molecular design. JoPMol integrates gene expression profiles, textual structural intentions, and property optimization within a unified diffusion-based modeling framework. By aligning multiple control signals, JoPMol enables controllable molecular generation and optimization. Extensive experiments show that JoPMol achieves consistently improved performance compared with baseline methods across multiple controllability and quality metrics under realistic evaluation settings.

Despite these results, JoPMol is currently evaluated primarily through in silico analyses and molecular docking, and the designed candidates have not yet been validated in wet-lab experiments. Future work will focus on extending experimental validation, expanding to broader disease targets and biological settings, and incorporating downstream biological screening to further examine the practical potential of jointly controlled precision molecular design in drug discovery.

\bibliographystyle{ACM-Reference-Format}
\bibliography{references}

\appendix
\newpage

\begin{center}
    {\Large\bf Appendix}
\end{center}
\setcounter{table}{0}
\section{Dataset Details}
\label{appendix_A}
To support heterogeneous conditional molecule generation scenarios, we construct three transcriptomic datasets corresponding to different biological supervision signals, referred to as Data~I, Data~II, and Data~III. All datasets are uniformly preprocessed by retaining the 978 landmark genes defined in the LINCS L1000 \cite{duan2014lincs} platform and applying feature-wise normalization to ensure numerical comparability across conditions. The raw sources of these datasets are publicly available under licenses permitting academic or research use.

\textbf{Data I} consists of compound-induced gene expression profiles derived from human cell lines under small-molecule perturbations. Each sample reflects a specific experimental setting characterized by the administered compound, dosage, exposure duration, and cellular context. In this study, we focus on profiles collected from a single representative cell line, resulting in a large-scale collection covering more than ten thousand unique compounds. These expression signatures provide supervision signals that capture how molecular interventions reshape transcriptional states, enabling the model to associate chemical structures with downstream cellular responses.

\textbf{Data II} contains target-level perturbation signatures generated through genetic modulation, including knockdown or overexpression of individual protein targets. Each profile encodes the transcriptomic consequence of manipulating a specific biological regulator. We select a subset of biologically meaningful targets spanning kinase signaling, epigenetic regulation, and cancer-related pathways. These data enable conditioning molecule generation on desired target-oriented functional effects, supporting controllable structure design driven by mechanistic intervention cues rather than purely chemical similarity.

\textbf{Data III} is composed of disease-associated transcriptomic signatures derived from patient-level expression studies. Each disease signature is obtained by aggregating multiple samples into a representative profile to reduce individual variability and enhance robustness. In this work, only gastric cancer–related profiles are included. This dataset allows the model to learn disease-conditioned molecular generation, aligning chemical design with pathological transcriptional patterns.


\section{Annotation Validation Protocol for Structural Design Intent}
\label{appendix_B}
\renewcommand{\thetable}{B.\arabic{table}} 
\renewcommand{\thefigure}{B.\arabic{figure}}

To ensure the reliability and consistency of the automatically extracted structural design intents, we adopt a multi-stage human validation protocol involving independent expert review, consensus adjudication, and quality auditing.

\paragraph{Automatic intent extraction.}
For each molecule, its SELFIES sequence is first processed by the BioT5 encoder to generate a candidate structural intent representation, which summarizes semantic patterns related to functional groups, substructures, and local chemical motifs. These automatically generated intents serve as the initial annotations.

\paragraph{Independent expert review.}
Two domain experts with formal training in medicinal chemistry independently examine each candidate intent without access to each other's judgments. Reviewers are provided with the original molecular structure (rendered from SELFIES), the predicted intent description, and a standardized checklist covering functional group correctness, structural completeness, chemical plausibility, and semantic consistency. Each intent is labeled as \textit{accepted}, \textit{minor revision}, or \textit{rejected}, and optional correction notes are recorded when necessary.

\paragraph{Consensus adjudication.}
All samples with conflicting labels or revision requests are jointly reviewed in a consensus meeting. Reviewers discuss discrepancies, reconcile interpretations, and produce a finalized intent annotation. If consensus cannot be reached, the sample is conservatively excluded from downstream training and evaluation to avoid introducing noisy supervision.

\paragraph{Quality auditing and spot checking.}
A random subset of finalized annotations is periodically re-evaluated to detect potential systematic bias or drift in annotation criteria. This auditing step ensures long-term consistency across the dataset and guards against latent labeling artifacts introduced during scaling.

\section{Baseline Models}

To comprehensively evaluate JoPMol, we select representative baseline models from three complementary perspectives: (i) gene expression–conditioned molecular generation, (ii) text-guided molecular generation, and (iii) chemically informed generative modeling. All baselines are configured following their official implementations whenever possible and evaluated under consistent preprocessing and benchmarking settings.

\paragraph{TRIOMRHE}
TRIOMRHE is a Transformer-based framework that learns the mapping between transcriptomic signatures and molecular responses through multi-head attention and heterogeneous embedding strategies. It models how gene expression patterns correspond to chemical structures at the representation level. We follow the official configuration and apply it to our benchmark data.

\paragraph{Gx2Mol.}
Gx2Mol is a hybrid neural architecture designed to generate molecules conditioned on transcriptomic signals. It combines multiple neural components to encode gene expression features and translate them into molecular representations. We apply the released implementation using our normalized gene profiles.

\paragraph{GxVAEs.}
GxVAEs formulates transcriptome-guided molecule generation under a conditional variational autoencoder framework, where gene expression vectors and molecular structures are coupled through a shared latent space. This enables the model to reconstruct or sample molecules aligned with specific transcriptional states. Pretrained weights are used for evaluation.

\paragraph{MolSearch.}
MolSearch is a retrieval-augmented molecular generation framework that formulates molecule discovery as a neural search problem over a learned chemical space. It iteratively proposes candidate molecules and refines them based on similarity and property-driven feedback, enabling efficient exploration of chemically valid regions. We adopt the official implementation and use MolSearch as a representative search-based baseline for property-oriented molecular optimization.

\paragraph{DyMol.}
DyMol is a dynamic molecular optimization framework that combines reinforcement learning with oracle-guided evaluation to iteratively improve molecular candidates under target property constraints. It employs adaptive policy updates to balance exploration and exploitation during optimization, enabling flexible control over multiple physicochemical objectives. We use the standard configuration and include DyMol as a representative reinforcement learning-based baseline for property-driven molecular optimization.

\paragraph{FRATTVAE}
FRATTVAE is a variational autoencoder model that integrates recurrent encoders with attention mechanisms for molecular sequence modeling. It learns latent molecular representations and reconstructs valid SMILES through attentive decoding, enabling controllable generation in latent space. We adopt the standard implementation and use it as a representative chemically driven generative baseline.

\paragraph{Mol-T5.}
Mol-T5 is a Transformer-based large language model pretrained on large-scale molecular SMILES corpora. We employ the mol-t5-base checkpoint and prompt the model directly with target-related textual descriptions to generate candidate molecules without additional fine-tuning.

\paragraph{Text+ChemT5.}
This baseline integrates natural language prompts with chemically grounded representations using ChemT5, which is pretrained jointly on textual and molecular modalities. Molecular descriptions and target cues are concatenated and provided as input, and generation is performed in a zero-shot setting.

\paragraph{TGM-DLM}
TGM-DLM is a two-stage diffusion framework for text-conditioned molecular generation. The first stage predicts a coarse molecular scaffold from textual input, while the second stage progressively refines it into a chemically valid molecule via conditional diffusion. Official implementations are adopted for fair comparison.

\section{Metric Details}
\label{appendix_D}
\renewcommand{\thetable}{D.\arabic{table}} 
\renewcommand{\thefigure}{D.\arabic{figure}}

To provide a comprehensive evaluation of JoPMol, we report three groups of metrics: (i) \emph{structure similarity} between generated and reference molecules (ECFP, APFP, Dice), (ii) \emph{sequence similarity} between their SELFIES strings (BLEU, Levenshtein, LCS), and (iii) \emph{overall statistics} that summarize generation quality at the set level (Diversity, Total, and Rank). These metrics jointly reflect structural fidelity, sequence-level consistency, and overall generation reliability.

\paragraph*{ECFP Similarity ($\uparrow$).}
We compute the Tanimoto similarity between ECFP (extended-connectivity fingerprint) of a generated molecule $g$ and its reference $r$:
\[
\text{ECFP}(g,r)=\frac{|\mathbf{f}_g\cap \mathbf{f}_r|}{|\mathbf{f}_g\cup \mathbf{f}_r|}.
\]
Higher values indicate closer local substructure agreement.

\paragraph*{APFP Similarity ($\uparrow$).}
APFP (atom-pair fingerprint) captures atom-type pairs and their topological distances, emphasizing more global structural relations. We compute the Tanimoto similarity in the same manner as ECFP, but on atom-pair fingerprints.

\paragraph*{Dice Similarity ($\uparrow$).}
Dice similarity measures fingerprint overlap using:
\[
\text{Dice}(g,r)=\frac{2|\mathbf{f}_g\cap \mathbf{f}_r|}{|\mathbf{f}_g|+|\mathbf{f}_r|},
\]
which is more sensitive to shared features.

\paragraph*{BLEU ($\uparrow$).}
BLEU evaluates $n$-gram overlap between the generated and reference SELFIES sequences, capturing local token-level agreement.

\paragraph*{Levenshtein ($\downarrow$).}
Levenshtein distance counts the minimum number of edit operations required to transform the generated SELFIES into the reference sequence. Lower values indicate better sequence consistency.

\paragraph*{LCS ($\uparrow$).}
LCS (Longest Common Subsequence) measures the length of the longest subsequence shared by two SELFIES strings while preserving order, reflecting long-range sequential consistency.

\paragraph*{Diversity ($\uparrow$).}
Diversity quantifies the heterogeneity of the generated set by measuring average pairwise dissimilarity in fingerprint space (higher is better), indicating broader chemical coverage and reduced mode collapse.

\paragraph*{Total Score ($\uparrow$).}
Following our benchmark protocol, the overall Total score is defined as the product of three set-level generation metrics:
\[
\text{Total}=\text{Validity}\times\text{Novelty}\times\text{Uniqueness}.
\]
This multiplicative form penalizes models that perform poorly in any single aspect and favors methods that simultaneously achieve reliable validity, sufficient novelty, and low redundancy.

\paragraph*{FCD ($\downarrow$).}
FCD (Fréchet ChemNet Distance) measures the distributional distance between generated molecules and reference molecules in the learned chemical feature space of a pretrained ChemNet model, reflecting the overall similarity of chemical properties and structural characteristics between the two molecule sets.

\paragraph*{Rank ($\downarrow$).}
To obtain a unified ranking across metrics, we first compute the per-metric rank of each model (with larger-is-better for $\uparrow$ metrics and smaller-is-better for $\downarrow$ metrics). The final Rank score is then defined as the average of these ranks over the selected metric set $\mathcal{M}$:
\[
\text{Rank}=\frac{1}{|\mathcal{M}|}\sum_{m\in\mathcal{M}}\operatorname{rank}_m,
\]
where $\operatorname{rank}_m$ denotes the ranking position under metric $m$ (ties, if any, are handled by assigning averaged ranks). A lower Rank indicates better overall performance.

\section{Implementation Details.}
\label{appendix_E}
\renewcommand{\thetable}{E.\arabic{table}} 
\renewcommand{\thefigure}{E.\arabic{figure}}
\subsection{Experiment Settings}

\begin{table*}[t]
\caption{
Effect of the loss weighting coefficient $\lambda$ in JoPMol.
Moderate weighting ($\lambda=0.3$) achieves the best trade-off between structural similarity and distributional consistency.
}
\label{table:metrics_l}
\centering
\resizebox{1.0\textwidth}{!}{
\renewcommand{\arraystretch}{1.2}
\begin{tabular}{l|ccccccccc}\toprule
Values & Validity(\%)$\uparrow$ & Uniqueness(\%)$\uparrow$ & Novelty(\%)$\uparrow$ & BLEU$\uparrow$ & FCD$\downarrow$ & Leven$\downarrow$ & Dice$\uparrow$ & APFP$\uparrow$ & ECFP$\uparrow$ \\\midrule
$\lambda=0.0$ & 100.0 & 99.49 & \textbf{100.0} & 0.43 & 15.03 & 28.84 & 0.27 & 0.31 & 0.15 \\
$\lambda=0.1$ & 100.0 & 98.33 & 99.85 & 0.43 & 15.84 & 28.98 & 0.25 & 0.31 & 0.15 \\
$\lambda=0.3$ & 100.0 & 98.91 & 94.29 & \textbf{0.61} & \textbf{7.40} & \textbf{19.34} & \textbf{0.46} & \textbf{0.51} & \textbf{0.33} \\
$\lambda=0.5$ & 100.0 & 99.42 & \textbf{100.0} & 0.42 & 16.57 & 29.51 & 0.25 & 0.30 & 0.14 \\
$\lambda=0.7$ & 100.0 & 99.49 & \textbf{100.0} & 0.45 & 16.32 & 29.50 & 0.25 & 0.31 & 0.15 \\
$\lambda=0.9$ & 100.0 & \textbf{99.78} & \textbf{100.0} & 0.45 & 15.75 & 29.47 & 0.25 & 0.31 & 0.15 \\
$\lambda=1.0$ & 100.0 & \textbf{99.78} & 99.93 & 0.44 & 15.53 & 29.51 & 0.26 & 0.31 & 0.15 \\\bottomrule
\end{tabular}
}
\end{table*}

All experiments are implemented in PyTorch with Python~3.8 and trained on a workstation equipped with an NVIDIA A30 GPU. Molecular structures are encoded as SELFIES sequences, with the maximum sequence length fixed at 256 tokens.

Gene expression profiles are projected through a stack of three fully connected layers with hidden dimensions of 512, 256,128 and 64, respectively, yielding a compact latent representation for conditional control. The same multilayer configuration is applied symmetrically during feature reconstruction. The optimization process adopts a learning rate of $1\times10^{-4}$ and applies dropout with a ratio of 0.2 to stabilize training.

For structural semantic conditioning, tokenized SELFIES sequences are embedded into a 32-dimensional trainable embedding space. In parallel, molecular textual descriptions are encoded using a frozen language encoder with a 768-dimensional hidden representation to extract high-level semantic features, which are subsequently aligned with structural intent representations.

The generative backbone follows a diffusion-based formulation with $T=2000$ denoising steps, optimized using a learning rate of $1\times10^{-4}$ and a dropout rate of 0.1. To further accelerate generation over remapped SELFIES sequences, a uniform skip-sampling strategy is applied during the reverse diffusion process. For the relevance alignment mechanism in JoPMol, the weighting coefficient $\lambda$ is fixed at 0.3 and used consistently in all experiments.

We further investigate the impact of the loss weighting coefficient $\lambda$, which controls the contribution of the relevance-guided objective in JoPMol.
As shown in Table~\ref{table:metrics_l}, the choice of $\lambda$ significantly influences the trade-off between structural alignment, distributional consistency, and diversity.
When $\lambda$ is small (e.g., $\lambda=0.0$ or $0.1$), the model relies primarily on the unconditional diffusion objective, resulting in high novelty and diversity but relatively weak structural alignment, as reflected by lower fingerprint similarities and BLEU scores.
Conversely, when $\lambda$ becomes large (e.g., $\lambda \geq 0.5$), the model is overly constrained by the relevance objective, which leads to degraded distributional consistency (higher FCD) and poorer sequence alignment (higher Levenshtein distance), despite maintaining high validity and novelty.
A moderate weighting ($\lambda=0.3$) achieves the best balance across multiple criteria.
Specifically, it significantly improves structural similarity metrics (Dice, APFP, and ECFP) and sequence-level alignment (BLEU), while also yielding the lowest FCD and Levenshtein distance.
These results indicate that an appropriate balance between conditional guidance and generative flexibility is crucial for achieving high-quality and controllable molecular generation.

\subsection{Use of Large Language Models}
We use a large language model (LLM) solely for language polishing and grammar refinement to improve the clarity and readability of the manuscript. The LLM is not involved in the generation of scientific content, experimental design, data analysis, model implementation, or result interpretation. All technical contributions, experiments, and conclusions are conducted and verified by the authors.

\section{Experimental Supplements}
\label{appendix_F}
\renewcommand{\thetable}{F.\arabic{table}} 
\renewcommand{\thefigure}{F.\arabic{figure}}

\subsection{Additional Evaluation}

\begin{table}[t]
\caption{Additional evaluation of JoPMol and baseline models on structural similarity, property consistency, and overall generation quality.}
\centering
\setlength{\tabcolsep}{3pt}
\resizebox{0.45\textwidth}{!}{
\begin{tabular}{l|cc|ccc}\toprule
\multirow{2}{*}{Method} & \multicolumn{2}{c|}{Structure similarity}  & \multicolumn{3}{c}{Overall statistics}\\
 & RDKit $\uparrow$ & MACCS $\uparrow$ & Validity
 $\uparrow$ & Unique $\uparrow$ & Novelty $\uparrow$ \\
\midrule
TRIOMPHE      & 0.31 & 0.39 & 50.75 & 78.68 & 87.57 \\
Gx2Mol        & 0.36 & 0.47 & 87.80 & 81.42 & 76.31 \\
GxVAEs        & 0.36 & 0.48 & 86.59 & 87.57 & 89.25 \\
Molsearch     & \textbf{0.55} & 0.60 & 100.0 & 93.01 & 88.10 \\
Dymol         & 0.23 & 0.35 & 100.0 & 90.23 & 91.23 \\
FRATTVAE      & 0.27 & 0.37 & 100.0 & 96.92 & 95.88 \\
Mol-T5        & 0.34 & 0.54 & 77.88 & 93.21 & 97.66 \\
Text+ChemT5   & 0.35 & 0.58 & 67.12 & 96.01 & \textbf{97.78} \\
TGM-DLM       & 0.37 & 0.54 & 82.33 & 93.74 & 91.73 \\
\midrule
JoPMol       & \underline{0.39} & \textbf{0.61} & \textbf{100.0} & \textbf{98.91} & 94.29 \\
\bottomrule
\end{tabular}
}

\label{tab:exp1_res}
\vspace*{-1.0\baselineskip} 
\end{table}

To further examine the robustness and practical utility of JoPMol, we conduct an additional evaluation using complementary metrics that explicitly measure structural similarity, property alignment, and overall generation quality. As summarized in Table \ref{tab:exp1_res}, the comparison covers representative transcriptome-guided models (TRIOMPHE, Gx2Mol, GxVAEs), chemically driven generative models (FRATTVAE), text-based generators (Mol-T5, Text+ChemT5), as well as a diffusion-based baseline (TGM-DLM).

\paragraph{Structural similarity.}
JoPMol achieves the strongest structural consistency across all evaluated fingerprint-based metrics. Specifically, JoPMol attains the highest RDKit Tanimoto similarity (0.39) and MACCS similarity (0.61), while simultaneously yielding the lowest Fr\'echet ChemNet Distance (7.40), indicating superior alignment between the generated molecular distribution and the reference set. Compared with the best competing baseline (TGM-DLM with FCD of 8.94 and MACCS of 0.54), JoPMol reduces distributional discrepancy by approximately 17\% and improves substructure overlap by more than 13\%, demonstrating its ability to preserve chemically meaningful structural patterns under multi-conditional guidance.


\paragraph{Overall generation quality.}
JoPMol maintains perfect validity (100.0\%) and the highest uniqueness (98.91\%) among all compared approaches, confirming stable decoding and low redundancy during generation. Although novelty is slightly lower than that of purely text-driven models, it remains at a competitive level (94.29\%), indicating that JoPMol balances exploration and fidelity rather than over-optimizing novelty at the cost of structural plausibility. Notably, several baselines exhibit trade-offs between validity and novelty, whereas JoPMol consistently preserves high-quality generation across all three criteria.

Overall, these results demonstrate that JoPMol achieves a favorable balance between structural fidelity, property alignment, and generative reliability. By jointly integrating transcriptomic signals, semantic structural intent, and chemical property constraints, the model effectively avoids the common failure modes observed in single-modality baselines, such as structural drift, property mismatch, or reduced validity. The supplementary evaluation therefore corroborates the primary experimental conclusions and highlights the practical advantages of multi-source controllable molecular generation.

\subsection{Gene Expression Reconstruction via VAE.}
\begin{figure}[t]
\centering
\includegraphics[width=1.0\linewidth]{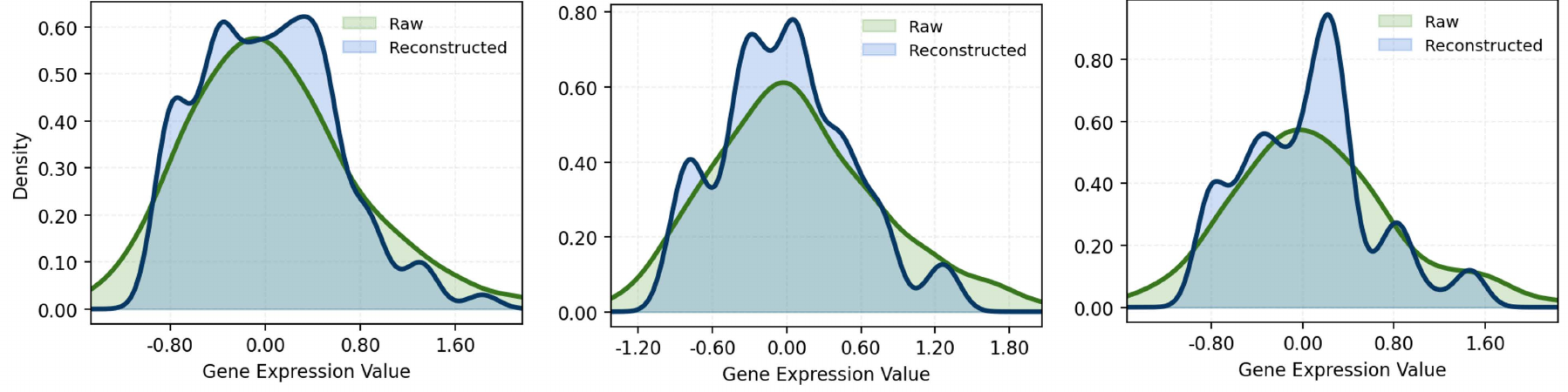}
\caption{Kernel density distributions of raw and reconstructed gene expression profiles for three randomly selected patients. Reconstruction is obtained by encoding the profiles into the latent space and decoding them through the VAE, demonstrating close distributional alignment.}
\label{fig:VAE}
\vspace*{-1.0\baselineskip} 
\end{figure}

To verify that the learned latent representations preserve informative transcriptomic patterns, we conduct a reconstruction-based validation using the gene expression encoder–decoder architecture. Specifically, three patient-level gene expression profiles are randomly sampled from the test set and projected into the latent space through the encoder. The corresponding latent embeddings are then decoded back into reconstructed gene expression profiles.

Figure \ref{fig:VAE} visualizes the empirical distributions of the original and reconstructed gene expression values for the three samples using kernel density estimation. Across all cases, the reconstructed distributions closely match the raw distributions in terms of central tendency, spread, and overall shape, indicating that the encoder effectively captures the dominant statistical characteristics of high-dimensional transcriptomic signals.

Notably, the reconstructed profiles preserve both the unimodal structure and the tail behavior of the original distributions, suggesting that the latent bottleneck does not excessively smooth or distort biologically meaningful variability. Minor deviations observed at extreme value ranges are expected due to stochastic sampling and regularization effects inherent to variational modeling.

These results provide qualitative evidence that the learned latent representations maintain sufficient fidelity for downstream conditional molecular generation, supporting their use as reliable conditioning signals in JoPMol.


\begin{table*}[t]
\caption{Comparison between numerical chemical property values (Value) and textual property descriptions (Text) as conditional inputs under the same experimental setting.}
\centering
\resizebox{1.0\textwidth}{!}{
\renewcommand{\arraystretch}{1.2}
\begin{tabular}{l|cccccccccccc}\toprule
Values & Validity$\uparrow$ & Uniqueness$\uparrow$ & Novelty$\uparrow$ & BLEU$\uparrow$ & FCD$\downarrow$ & Leven$\downarrow$ & MACCS$\uparrow$ & RDK$\uparrow$ & ECFP $\uparrow$ & logP& SA & QED \\\midrule
Text & 100.0 & \textbf{99.13} & \textbf{92.53} & 0.53 & 12.22 & \textbf{21.52} & 0.51 & 0.33 & 0.23 & 3.05 & 3.76 & 0.61\\

Value & 100.0 & 98.69 & 91.55 & \textbf{0.54} & \textbf{11.95} & 21.64 & \textbf{0.52} & \textbf{0.34} & 0.23 & \textbf{2.71} & \textbf{3.73} & \textbf{0.62} \\\bottomrule
\end{tabular}
}

\label{table:F3}
\vspace*{-1.0\baselineskip} 
\end{table*}

\begin{figure}[t]
\centering
\includegraphics[width=1.0\linewidth]{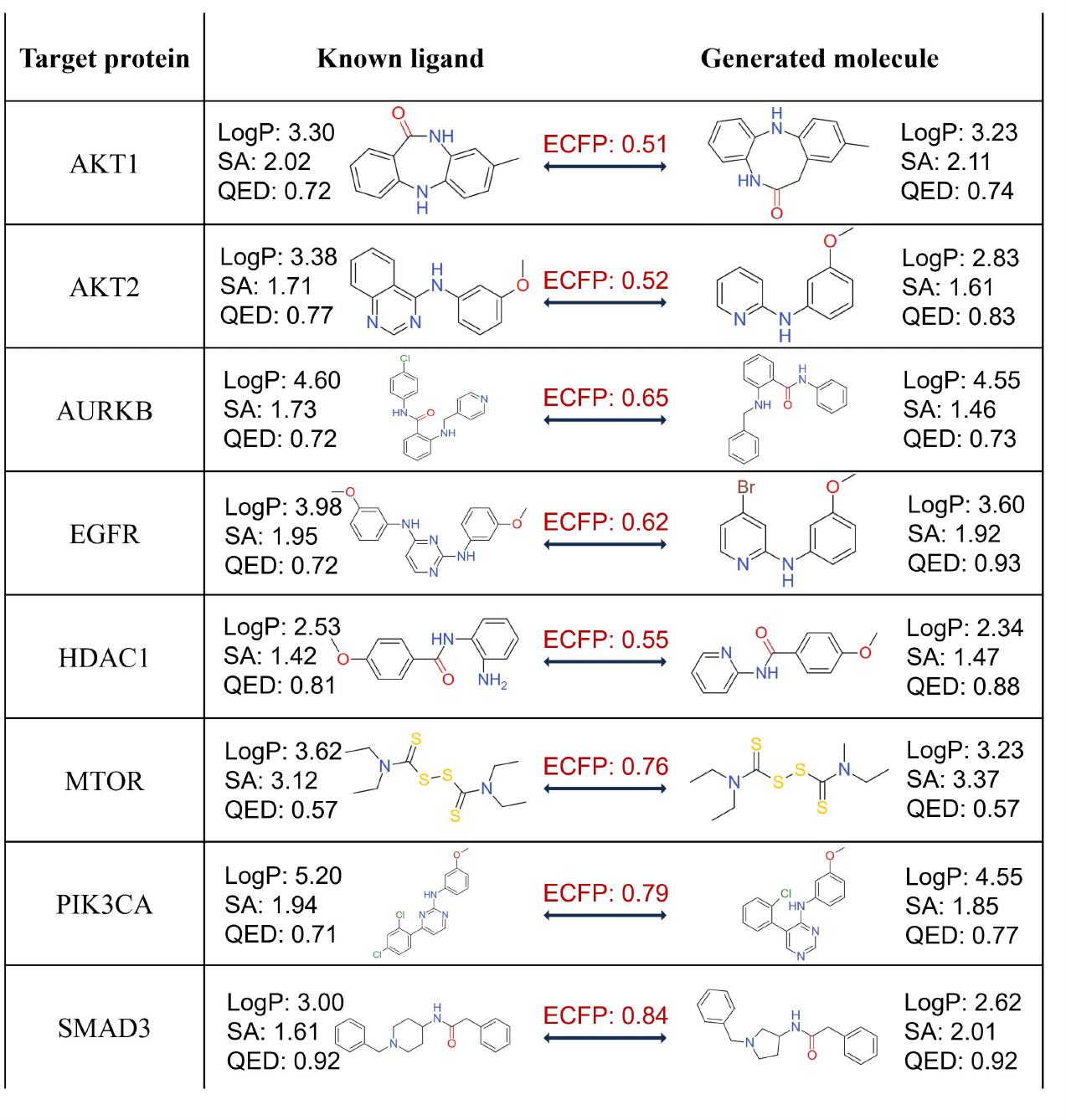}
\caption{Qualitative comparison between known ligands and generated molecules for eight additional targets (AKT1, AKT2, AURKB, EGFR, HDAC1, MTOR, PIK3CA, and SMAD3).}
\label{fig:additional_ligands}
\vspace*{-1.0\baselineskip} 
\end{figure}

\subsection{Chemical Property Representation Analysis}

Table \ref{table:F3} compares the performance of using numerical chemical property values (Value) versus textual property descriptions (Text) as conditional inputs. Both settings achieve similarly high validity and uniqueness, indicating comparable basic generation feasibility. However, the Value setting consistently exhibits more stable performance on structure- and distribution-related metrics. In particular, it achieves a lower FCD score than the Text setting (11.95 vs. 12.22), suggesting closer alignment with the target distribution. Comparable or slightly improved results are also observed on fingerprint similarity metrics, including MACCS, RDK, and ECFP. In addition, the Value setting achieves a higher QED score (0.62 vs. 0.61), reflecting more reliable preservation of drug-likeness.

In contrast, although the Text setting shows marginally better BLEU performance, this improvement does not translate into consistent gains in structural or property-level quality and may introduce additional semantic ambiguity and noise. Overall, directly conditioning on numerical chemical properties provides more precise, controllable, and stable optimization signals, and is therefore adopted in our model design.

\subsection{Additional Cross-Ligand Results}

Figure~\ref{fig:additional_ligands} further reports qualitative generation results for the remaining eight ligand conditions, including AKT1, AKT2, AURKB, EGFR, HDAC1, MTOR, PIK3CA, and SMAD3. For each target, the model receives transcriptomic perturbation profiles together with design intents as conditioning signals, and generates candidate molecules that are compared against the corresponding reference ligands in terms of structural similarity and physicochemical consistency.

Across all eight cases, JoPMol consistently preserves core scaffold patterns and key functional groups, yielding moderate to high structural similarity measured by ECFP (ranging approximately from 0.51 to 0.84). Targets such as MTOR, PIK3CA, and SMAD3 exhibit particularly strong structural transferability, reflecting the model’s ability to capture conserved substructure motifs under distinct biological perturbations. Even for more challenging targets with lower similarity scores (e.g., AKT1 and AKT2), the generated molecules still maintain chemically meaningful backbone alignment rather than degenerate or trivial structures.

In addition to structural fidelity, the generated molecules remain well aligned with the reference ligands in major physicochemical properties. The deviations in LogP, synthetic accessibility (SA), and drug-likeness (QED) are generally small across all targets, indicating stable control over lipophilicity, synthetic feasibility, and overall molecular quality. Notably, JoPMol avoids excessive property drift even when moderate structural variations are introduced, suggesting a balanced trade-off between transfer flexibility and property preservation.

Overall, these supplementary cases demonstrate that JoPMol maintains reliable cross-ligand generalization and transfer stability beyond the examples reported in the main text. The model consistently generates chemically plausible candidates that jointly satisfy biological conditioning, structural coherence, and property constraints, supporting its robustness for precision molecular design in diverse target settings.

\end{document}